\documentclass[10pt,twocolumn,letterpaper]{article}

\usepackage[final]{cvpr22}      

\makeatletter
\@namedef{ver@everyshi.sty}{}
\makeatother

\usepackage{graphicx}
\usepackage{amsmath}
\usepackage{amssymb}
\usepackage{booktabs}
\usepackage{pgffor}
\usepackage{multirow}

\usepackage[pagebackref,breaklinks,colorlinks]{hyperref}

\def\cvprPaperID{4260} 
\def\confName{CVPR}
\def\confYear{2022}

\begin{document}

\title{Low-light Image Enhancement via Breaking Down the Darkness}

\author{Qiming Hu, Xiaojie Guo\thanks{Corresponding Author}\\
College of Intelligence and Computing, Tianjin University, Tianjin 300350, China\\
{\tt\small huqiming@tju.edu.cn   xj.max.guo@gmail.com} 
}
\maketitle
\begin{abstract}
Images captured in low-light environment often suffer from complex degradation. Simply adjusting light would inevitably result in burst of hidden noise and color distortion. To seek results with satisfied lighting, cleanliness, and realism from degraded inputs, this paper presents a novel framework inspired by the divide-and-rule principle, greatly alleviating the degradation entanglement. Assuming that an image can be decomposed into texture (with possible noise) and color components, one can specifically execute noise removal and color correction along with light adjustment. Towards this purpose, we propose to convert an image from the RGB space into a luminance-chrominance one. An adjustable noise suppression network is designed to 
eliminate noise in the brightened luminance, having the illumination map estimated to indicate noise boosting levels. The enhanced luminance further serves as guidance for the chrominance mapper to generate realistic colors. Extensive experiments are conducted to reveal the effectiveness of our design, and demonstrate its superiority over state-of-the-art alternatives both quantitatively and qualitatively on several benchmark datasets. Our code is publicly available at \emph{\href{https://github.com/mingcv/Bread}{\emph{https://github.com/mingcv/Bread}}}.

\end{abstract}

\begin{figure}[t]
\begin{subfigure}{0.32\linewidth}
    \includegraphics[width=1\linewidth]{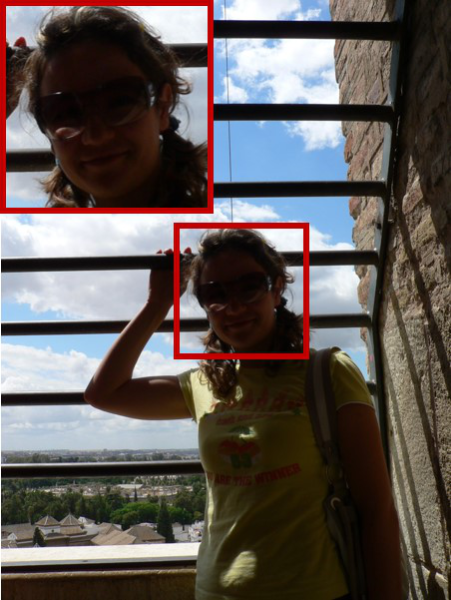}
    \subcaption*{Input}
\end{subfigure}
\begin{subfigure}{0.32\linewidth}
    \includegraphics[width=1\linewidth]{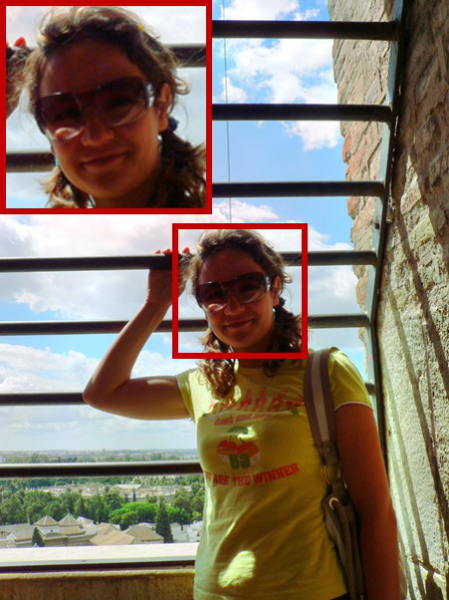}
    \subcaption*{DeepUPE \cite{wang2019underexposed}}
\end{subfigure}
\begin{subfigure}{0.32\linewidth}
    \includegraphics[width=1\linewidth]{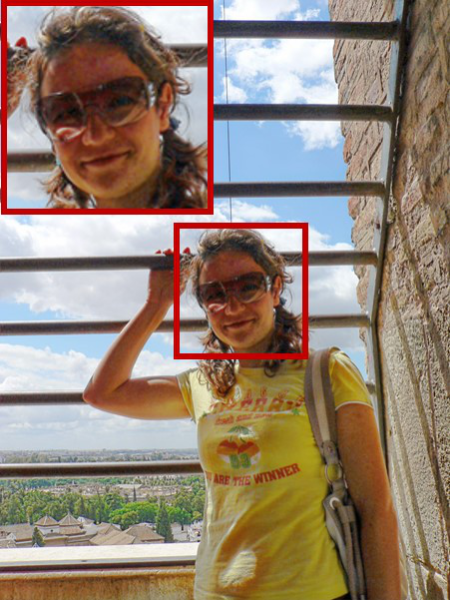}
    \subcaption*{KinD++ \cite{zhang2021beyond}}
\end{subfigure}

\begin{subfigure}{0.32\linewidth}
    \includegraphics[width=1\linewidth]{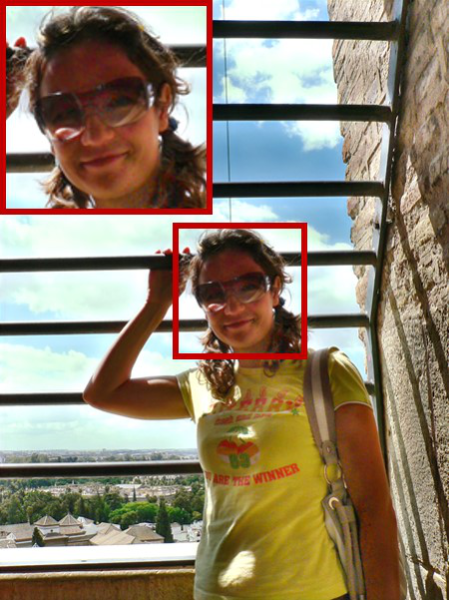}
    \subcaption*{DRBL \cite{yang2020fidelity}}
\end{subfigure}
\begin{subfigure}{0.32\linewidth}
    \includegraphics[width=1\linewidth]{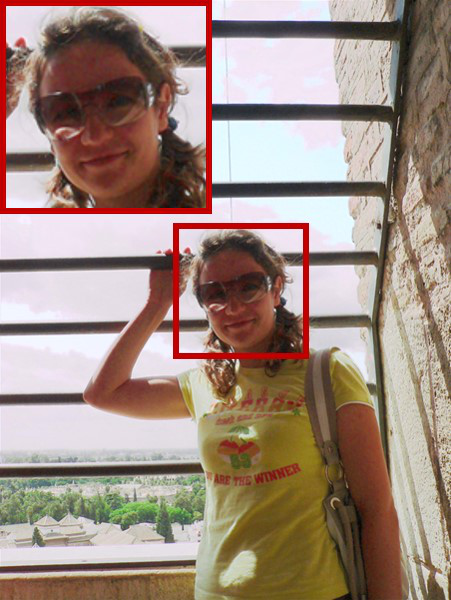}
    \subcaption*{DLN \cite{wang2020lightening}}
\end{subfigure}
\begin{subfigure}{0.32\linewidth}
    \includegraphics[width=1\linewidth]{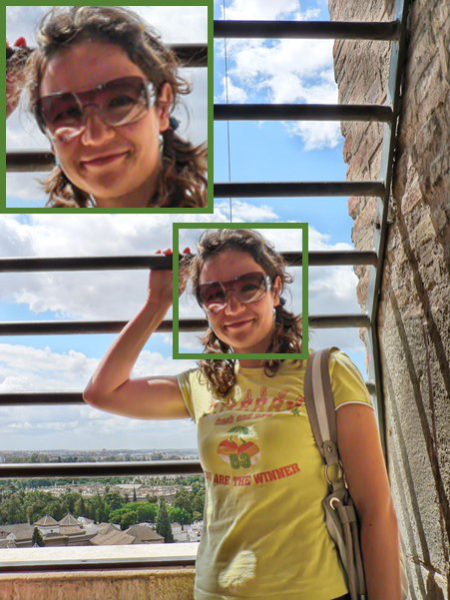}
    \subcaption*{Ours}
\end{subfigure}

\caption{Visual comparison on a sample from the VV dataset. Our method obtains striking improvement over the other competitors, \eg, the sky tone and the realism of human skin.} 
\label{fig:tease}
\end{figure}

\section{Introduction}
\label{sec:introduction}
  Capturing high-quality images under less controlled conditions is challenging especially using mobile devices. Often, people are shooting images in unsatisfactory light environment. For instance, we might take a photo against light source (please see Fig. \ref{fig:tease}); or a surveillance camera may be monitoring a place in the nighttime. In such cases, the images will suffer from poor visibility. To obtain high-quality images, a few solutions can be applied like one can extend exposure time to receive more information, but if the target scene is dynamic, the blur effect likely appears in captured images; another possible way is to set a flash for light compensation, which however frequently introduces unexpected highlights and unbalanced lighting into photos. Hence, instead of upgrading hardware, developing effective low-light enhancement techniques is highly desired for practical use. 

\emph{Low-light image enhancement is not a solo problem of light adjustment, which also has troubles of noise burst and color distortion concealed in the darkness because of the limited capability of photographing devices.} A number of methods follow the Retinex theory \cite{land1977retinex} through decomposing an image $I$ into its reflectance $R$ and illumination $L$ in the form of $I=R\circ L$, where $\circ$ designates the Hadamard product operator. Because the reflectance component reflects the intrinsic property of material, it is constant against variant illuminations. Ideally, once the \textbf{illumination} is estimated or given, the reflectance can be immediately obtained. One can adjust illumination according to different demands. However, due to limited quality of sensors, the \textbf{noise} factor $N$ will be always an annoying resident in images, thus the model turns out to be $I = R \circ L + N$. A simple algebraic transformation leads to $I=(R+\tilde{N})\circ L$, where $\tilde{N}=N/L$ with element-wise division. We can see from above that the noise will be also amplified along with light enhancement, which becomes spatially-correlated with $L$ as discussed in \cite{zhang2019kindling}. Furthermore, in low-light conditions, sensors (either CCD or CMOS) are sensitive and non-linear to insufficient photons of different light spectrum, which brings \textbf{color distortion} even with illumination and noise being properly handled. Therefore, for the sake of producing high-quality results from degraded inputs, a qualified algorithm should remedy the highly entangled illness of dim light, noise, and color distortion, simultaneously.

\begin{figure}[t]
\begin{minipage}{\linewidth}
\begin{subfigure}{0.32\linewidth}
    \includegraphics[width=1\linewidth]{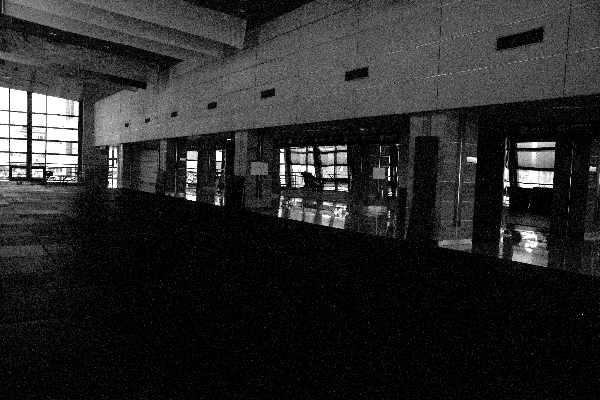}
    \subcaption*{R}
\end{subfigure}
\begin{subfigure}{0.32\linewidth}
    \includegraphics[width=1\linewidth]{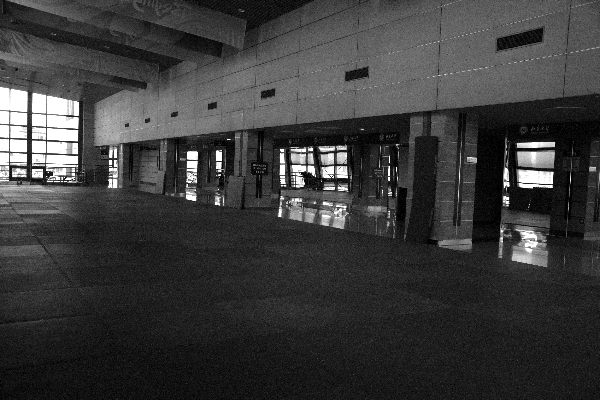}
    \subcaption*{G}
\end{subfigure}
\begin{subfigure}{0.32\linewidth}
    \includegraphics[width=1\linewidth]{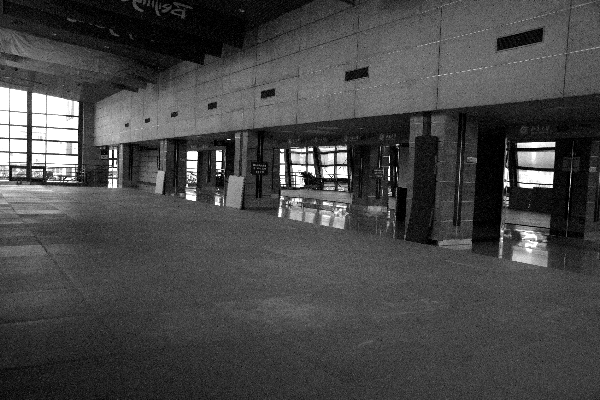}
    \subcaption*{B}
\end{subfigure}

\begin{subfigure}{0.32\linewidth}
    \includegraphics[width=1\linewidth]{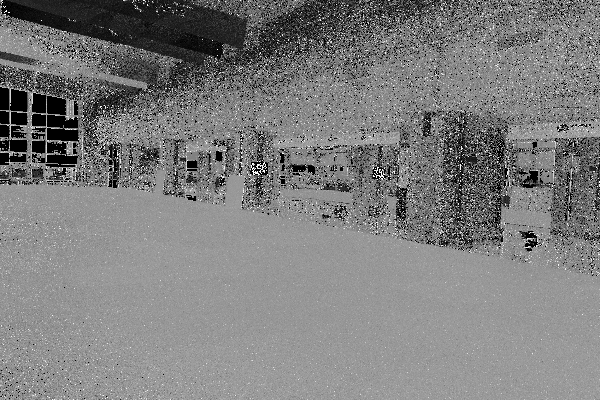}
    \subcaption*{H}
\end{subfigure}
\begin{subfigure}{0.32\linewidth}
    \includegraphics[width=1\linewidth]{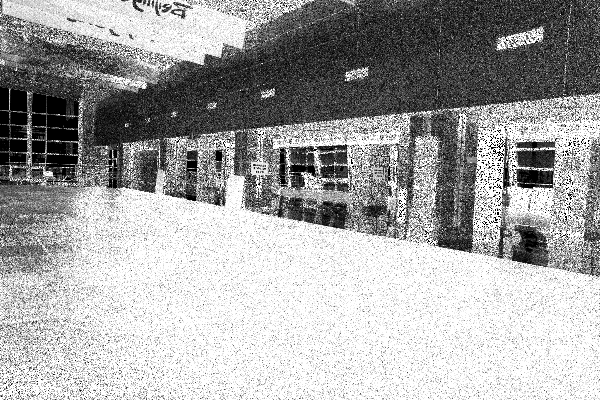}
    \subcaption*{S}
\end{subfigure}
\begin{subfigure}{0.32\linewidth}
    \includegraphics[width=1\linewidth]{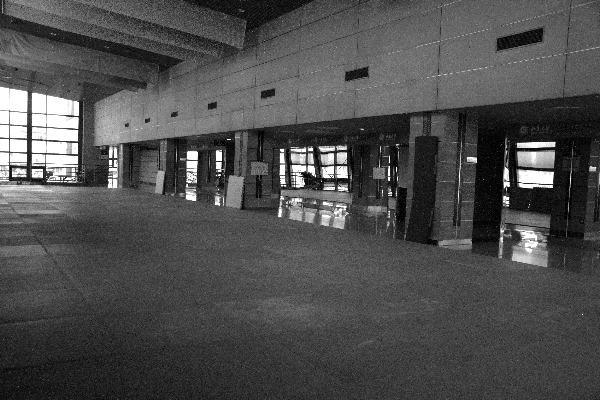}
    \subcaption*{V}
\end{subfigure}

\begin{subfigure}{0.32\linewidth}
    \includegraphics[width=1\linewidth]{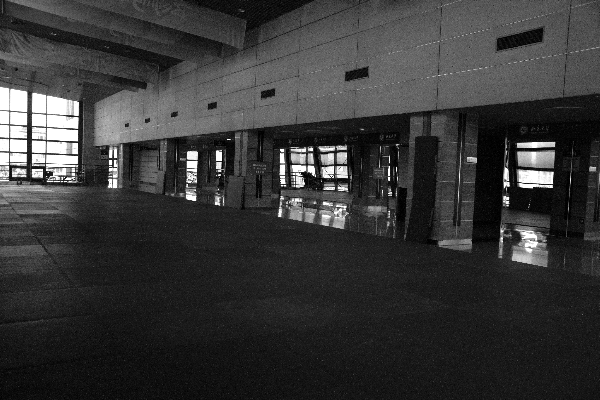}
    \subcaption*{Y}
\end{subfigure}
\begin{subfigure}{0.32\linewidth}
    \includegraphics[width=1\linewidth]{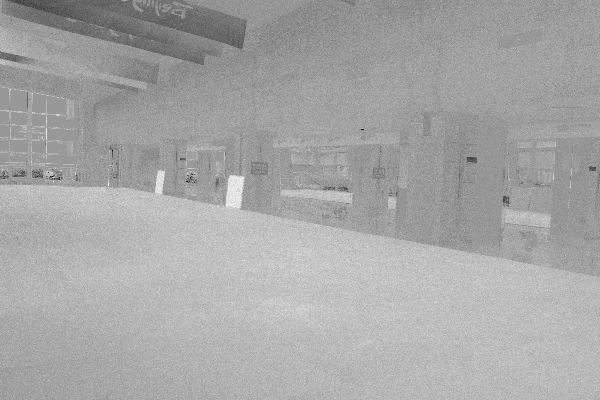}
    \subcaption*{Cb}
\end{subfigure}
\begin{subfigure}{0.32\linewidth}
    \includegraphics[width=1\linewidth]{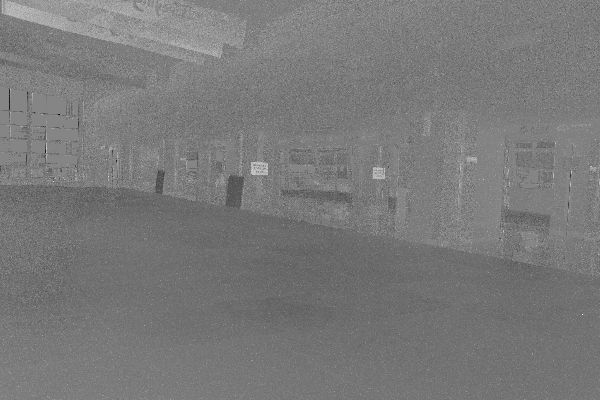}
    \subcaption*{Cr}
\end{subfigure}

\caption{An image from the LOL dataset, which is in low-light before being linearly enhanced. As can be observed in its RGB, HSV and YCbCr channels, the chrominance components in the YCbCr space are obviously less affected by noise than others.} 
\label{fig:colorspace}
\end{minipage}

\vspace{3pt}
\begin{minipage}{\linewidth}
\begin{subfigure}{0.49\linewidth}
    \includegraphics[width=1\linewidth]{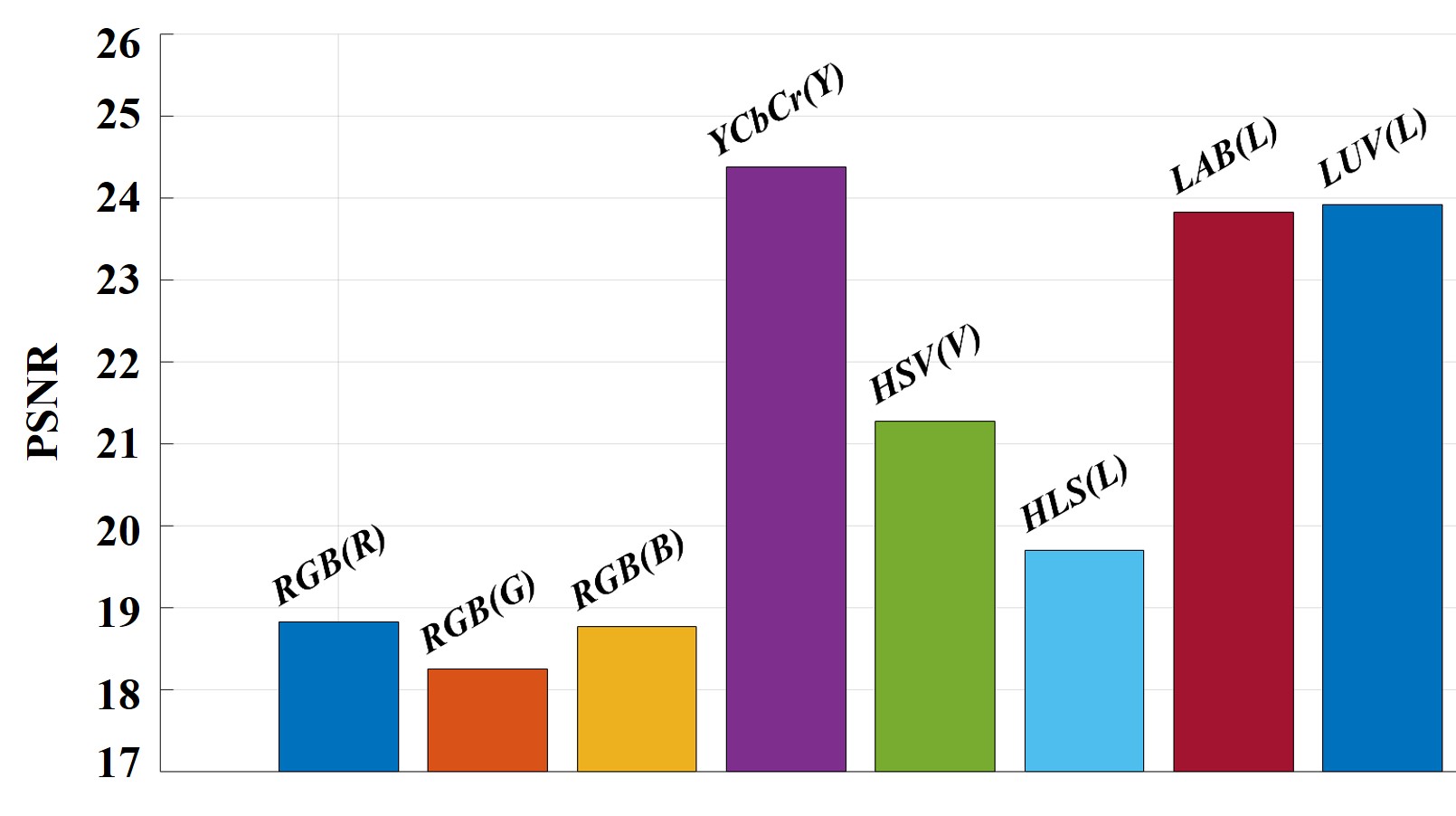}
    \subcaption*{Average PSNR}
\end{subfigure}
\begin{subfigure}{0.49\linewidth}
    \includegraphics[width=1\linewidth]{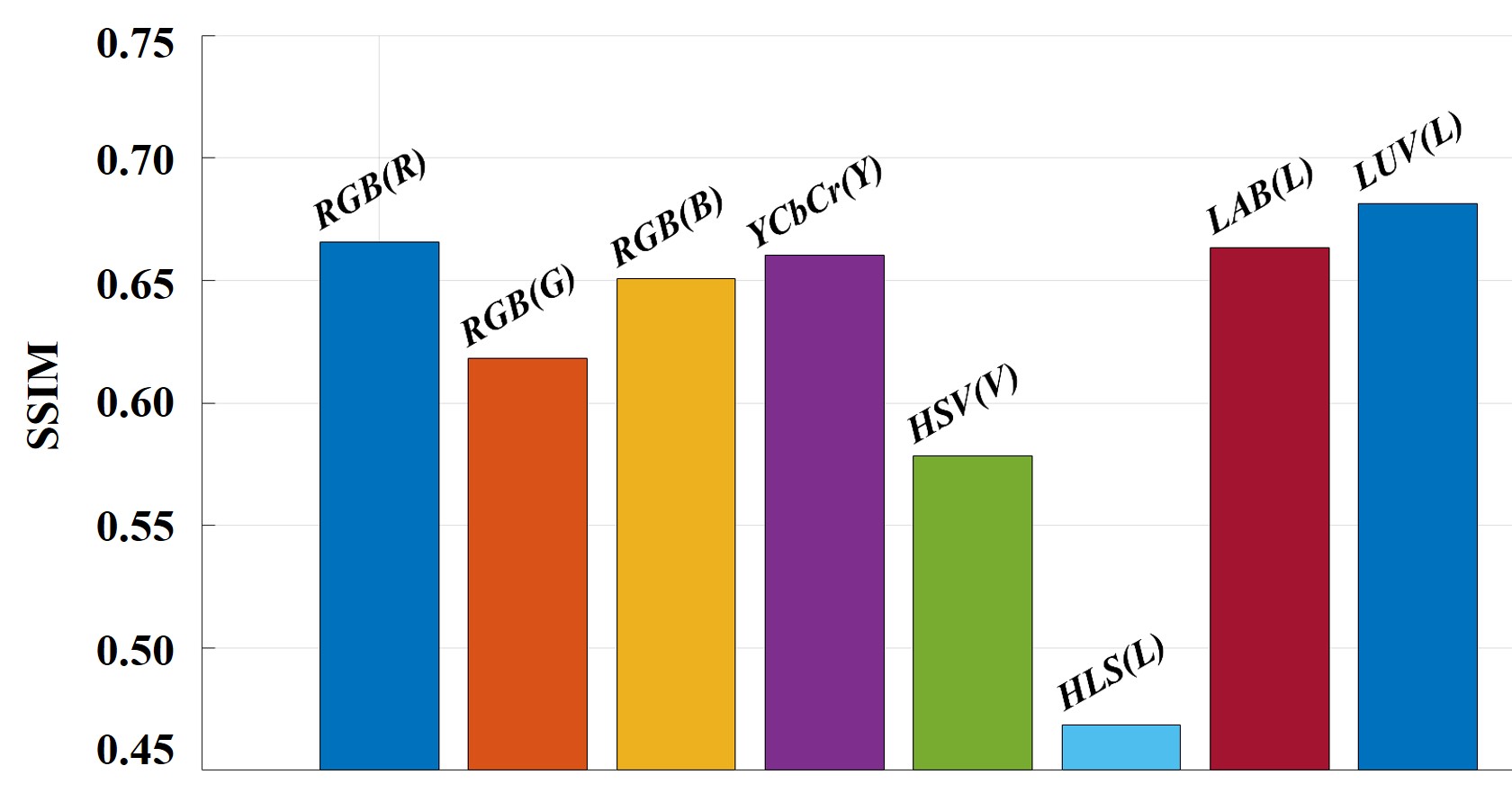}
    \subcaption*{Average SSIM}
\end{subfigure} 
\caption{The illuminations of low-light images in the LOL training dataset are first aligned  with their references and converted to different color-spaces. We replace one of their channels with the ground-truth one, and then compare them with references, obtaining average metrics that indicate to what extent a single channel in different spaces represents the major information of an image. }
\label{fig:colorspace_stat}
\end{minipage}
\end{figure}

This work studies how to handle the complex degradation in the darkness from a dividing and ruling perspective. Assuming that an image can be disassembled into texture (together with the main body of noise) and color components, the operations including noise removal and color correction along with light adjustment could be executed specifically. In Figs. \ref{fig:colorspace} and \ref{fig:colorspace_stat}, we visualize how the noise distributes in different color-spaces and make statistics on the effect of the single-channel restoration on data pairs from the LOL training dataset \cite{DBLP:conf/bmvc/WeiWY018}. The results suggest that the YCbCr space is a ``good" candidate to do the job of texture-color decomposition. 
As aforementioned, the principle behind our design is that \emph{``mind your own business"}. In other words, a solver is customized for coping with one type of degradation intently. By the luminance-chrominance separation, we are thereby possible to treat the degradation in the texture and color components individually, say focusing on the noise issue in the texture and the color distortion in the other part. Please notice that, although several methods \cite{yang2020fidelity,zhang2021beyond} have been recently proposed to take care of noise and color shift issues along with light enhancement without distinction on the recovered reflectance component, they barely consider decomposing the degradation from a texture-color point of view to further ease the problem.

This paper develops an effective low-light image enhancement framework via breaking down the darkness (Bread for short) based on the above analysis. The major contributions of this paper are summarized as follows:
\begin{enumerate}
    \item To the best of our knowledge, this is a pioneering attempt to decouple the entanglement of noise and color distortion, further mitigating the difficulty of low-light enhancement with complex degradation.
    \item We present an effective noise synthesis strategy under the guidance of illumination, significantly improving the suppression quality of amplified and spatially-correlated noise in the luminance.
    \item To tackle the color distortion issue left in light-enhanced images, we design a novel color adaption network, which can properly deal with the chrominance according to given luminance.
    \item Extensive comparisons together with ablation studies are provided to verify the efficacy of our method, and reveal its advance over other state-of-the-art methods both qualitatively and quantitatively. 
\end{enumerate}

\section{Related Work}
Numerous low-light image enhancement methods have been proposed over last decades,  which can be roughly grouped into traditional and deep learning-based methods.

\textbf{Traditional Methods.}
The simplest and most intuitive way is to linearly adjust the value range or execute a non-linear Gamma correction on inputs. Further, global and local histogram-based methods \cite{pisano1998contrast,cheng2004simple,abdullah2007dynamic,celik2011contextual,lee2013contrast} are introduced to expand the dynamic range of images. \emph{In spite of their ease of use, the enhancement quality is hardly guaranteed, due to the content-blindness.} Derived from the Retinex theory \cite{land1977retinex}, Single-scale Retinex (SSR) \cite{jobson1997properties} first uses the Gaussian blurred input as its illumination map, and then removes the estimated illumination from the input as its final result. Multi-scale Retinex (MSR) \cite{jobson1997multiscale} extends SSR by fusing the results of multiple Gaussian blur functions with different variances. Besides the above mentioned attempts, NPE \cite{wang2013naturalness} takes local maxima assumption to predict the illumination, which is manipulated by a mapper to act as the enhanced illumination and 
merged with the reflectance component. LIME \cite{guo2016lime} proposes to refine the initial illumination obtained by the Max-RGB assumption and structure preserving constraint. Although these methods can somewhat brighten low-light images, they barely take other hidden degradation, like noise and color distortion, into consideration. To suppress the noise effect, SRIE \cite{fu2016weighted} further imposes the sparsity on the recovered reflectance. Similarly, RRM \cite{li2018structure} integrates noise estimation into the main driving optimization to eliminate the noise in reflectance. However, \emph{the applicability of these optimization-based methods is limited because of the expensive computation, sensitivity to hyper-parameters, and unsatisfied enhancement quality.}

\textbf{Deep Learning-based Methods.}
Recently, methods based on deep learning have dominated the target task. For instance, MSR-Net \cite{shen2017msr} integrates the MSR mechanism into a neural network and uses the BM3D \cite{Dabov2007ImageDB} for denoising. It gains improvement in visual quality, but still suffers from the drawbacks like over-enhancement as the traditional MSR and lacks an embedded denoising functionality. LLNet \cite{lore2017llnet} synthesizes training pairs by randomly applying Gamma adjustment and adding synthetic noise to clean images. An auto-encoder network is then employed to learn the mapping function. However, the relationship between real-world illumination and noise is not touched, thus residual noise and over-smoothing problems show up. Cai \emph{et al.} \cite{cai2018learning} exploit to construct references from multi-exposure sequences for single image enhancement. However, its performance is upper-bounded by involved MEF methods. DUPE \cite{wang2019underexposed} and GLAD \cite{wang2018gladnet} learn the illumination map for image retouching. Despite reasonable results, they cannot effectively alleviate noise and color distortion issues. DRD \cite{DBLP:conf/bmvc/WeiWY018} and KinD++ \cite{zhang2021beyond} resort to the layer decomposition strategy that is beneficial to both illumination adjustment and reflectance refinement. DRBL \cite{yang2020fidelity} develops a deep recursive band network for semi-supervised low-light enhancement. DLN \cite{wang2020lightening} introduces lighten-darken trade-off and feature aggregation blocks to ameliorate results. However, they frequently have troubles in over-exposure and color distortion. In unsupervised settings, EnlightenGAN \cite{jiang2021enlightengan} attempts to take advantage of larger-scale unpaired training data through employing the GAN mechanism, while Zero-DCE \cite{guo2020zero} alternatively learns a set of non-reference loss functions. Although relieving the requirement of paired data, they mainly focus on the light factor, and thus have insufficient abilities to remedy other defects. \emph{Despite a progress made toward addressing the problem, effective and efficient designs for handling complex and multi-entangled degradation are desirable for practical use.}


\begin{figure*}[t]
\includegraphics[width=1\linewidth]{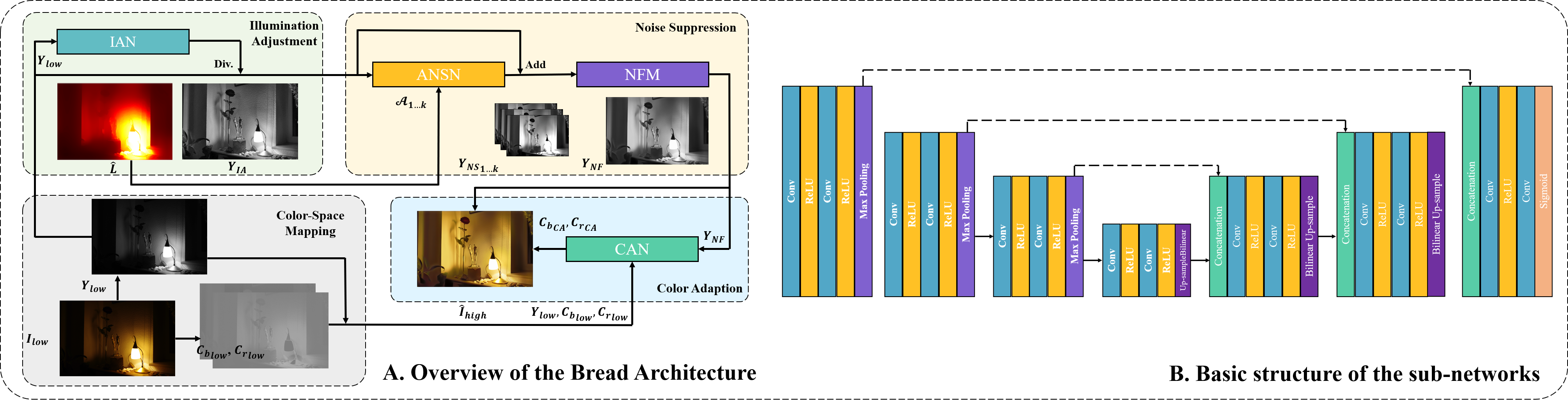}
\caption{An overview of our proposed Bread architecture and the basic structures of the involved sub-networks. The estimated relative illumination map are shown by heat map for a better view. } 
\label{fig:full_architecture}
\end{figure*}

\section{Methodology}
\subsection{Problem Analysis \& Motivation}
\label{sec:motivation}
As previously discussed, low-light image enhancement encounters complex degradation mixed up by dim light, noise, and color distortion. Most previous methods tend to directly restore the reflectance from illumination-adjusted input under the guidance of ground-truth data, which in nature ignores the noise-amplification process by the illumination adjustment operation. In other words, the defects hidden in the darkness will burst in the RGB channels of the reflectance map, which are correlated to the spatially-variant illumination. This fact is tricky for both end-to-end networks like \cite{wang2018gladnet} and post-processing methods like \cite{shen2017msr}, resulting in artifacts in texture and/or color. Motivated by the principle of divide-and-rule, we propose to convert images from the RGB colorspace into a luminance-chrominance one for decoupling the light and noise factors from the color degradation. Among various colorspaces, YCbCr seems to be a good choice supported by the evidence given in Fig. \ref{fig:colorspace_stat}. To handle the dim light and noise in the luminance, we denote the low-light luminance as $Y_{low} = Y\circ L_{low} + N$, and the normal-light reference as $Y_{high} = Y\circ L_{high}$. The connection between $Y_{low}$ and $Y_{high}$ can be found as follows:
\begin{equation}
     Y_{high} = \frac{(Y_{low}-N)\circ L_{high}}{{L}_{low}}=\frac{Y_{low}}{\hat{L}}-\frac{N}{\hat{L}},
     \label{eq:y_high}
\end{equation}
where $\hat{L} = L_{low}/L_{high}$ represents the relative difference between illuminations of low and normal-light images, and the division is element-wise. We propose a solver, \ie illumination adjustment network, to seek $Y_{IA} = Y_{low}/\hat{L}$. From Eq. \eqref{eq:y_high}, we can see that the noise term $\tilde{N} = N/\hat{L}$ from $Y_{low}$ remains in $Y_{IA}$. Even if $N$ is simple, $\tilde{N}$ will become much more complicated due to the correlation with spatially-variant illumination. Thus, it is natural to adopt $\hat{L}$ as an indicator for the denoising. Barely previous methods take into account this property. KinD \cite{zhang2019kindling} is limited to simply concatenate $L_{low}$ and $R$ together to restore the reflectance, which may not sufficiently exploit the guidance information from the illumination, and the single restoration network has to deal with all the degradation simultaneously. Inspired by the above, we alternatively synthesize noisy images that have the same illumination with references but are corrupted with stimulated amplified noise guided by $\mathcal{A} = \Phi(\hat{L})$, where $\Phi(\cdot)$ is a function that reflects the relationship between illuminations and noise levels. By this means, we obtain a noise-suppression solver, the solution of which are constrained by $\mathcal{A}$. Though the relative noise level map is determined, its overall scale is inaccessible due to the differences in photographing devices. To robustly remove the noise, we further fuse the denoised luminance $Y_{NS}$ under different suppression strengths to obtain the expected luminance map $Y_{NF}$. Having $Y_{NF}$ estimated, it is reasonable to employ it as guidance for mapping chrominance (color correction) from the input to an adjusted/target light levels, which is achieved by our color adaption network.

\subsection{Overall Network Design}

\begin{figure*}[t]
\foreach \mt/\app in {Input_plot/Input, LIME/LIME\,\cite{guo2016lime}, NPE/NPE\,\cite{wang2013naturalness}, DUPE/DUPE\,\cite{wang2019underexposed}, GLAD/GLAD\,\cite{wang2018gladnet}, DRBL/DRBL\,\cite{yang2020fidelity}, kindlev2/KinD++\,\cite{zhang2019kindling}, EnlightenGAN/EG\,\cite{jiang2021enlightengan}, ZeroDCE/Zero-DCE\,\cite{guo2020zero}, Bread/Bread, BreadMEF/Bread-ME}{
	\begin{subfigure}{0.15\linewidth}
        \includegraphics[width=1\linewidth]{figures/eval15_sota/665_\mt.png}
        \subcaption*{\app}
	\end{subfigure}
	\hfill
 }	
\begin{subfigure}{0.15\linewidth}
    \includegraphics[width=1\linewidth]{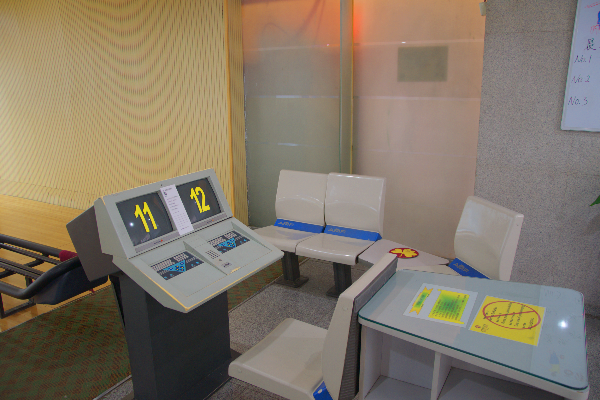}
    \subcaption*{GT}
\end{subfigure}

\foreach \mt/\app in {Input_plot/Input, LIME/LIME\,\cite{guo2016lime}, NPE/NPE\,\cite{wang2013naturalness}, DUPE/DUPE\,\cite{wang2019underexposed}, GLAD/GLAD\,\cite{wang2018gladnet}, DRBL/DRBL\,\cite{yang2020fidelity}, kindlev2/KinD++\,\cite{zhang2021beyond}, EnlightenGAN/EG\,\cite{jiang2021enlightengan}, ZeroDCE/Zero-DCE\,\cite{guo2020zero}, Bread/Bread, BreadMEF/Bread-ME}{
	\begin{subfigure}{0.15\linewidth}
        \includegraphics[width=1\linewidth]{figures/eval15_sota/79_\mt.png}
        \subcaption*{\app}
	\end{subfigure}
	\hfill
 }	
\begin{subfigure}{0.15\linewidth}
    \includegraphics[width=1\linewidth]{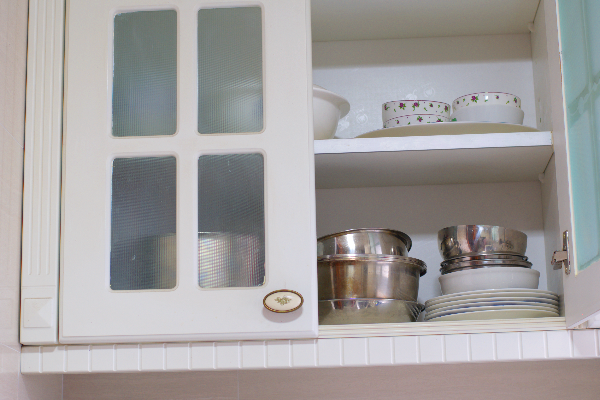}
    \subcaption*{GT}
\end{subfigure}

\caption{Visual comparison of state-of-the-art methods and ours on samples from the LOL dataset.} 
\label{fig:visual_comparsion_eval}
\end{figure*}
Figure \ref{fig:full_architecture} shows the overall architecture of our method, which comprises an illumination adjustment net (IAN), an adaptive noise suppression net (ANSN), and a color adaption net (CAN). As can be seen from Fig. \ref{fig:full_architecture} A, we firstly convert the input from the RGB to the YCbCr colorspace, obtaining the luminance $Y_{low}$ and chrominance $C_{b_{low}}, C_{r_{low}}$ components. Then, $Y_{low}$ is fed into the IAN to predict the relative illumination difference map $\hat{L}$, yielding the adjusted $Y_{IA}$. The ANSN takes in $Y_{IA}$ and $\Phi(\hat{L})$ (denoted as $\mathcal{A}$) to produce $Y_{NS}$. As discussed in \ref{sec:motivation}, we introduce a noise fusion module (NFM) into the noise suppression stage. Relative noise level maps $\{\mathcal{A}_{1},...,\mathcal{A}_{k}\}$ with $k$ different suppression strengths are used to generate $k$ denoised luminance maps $\{Y_{NS_{1}},...,Y_{NS_{k}}\}$, which are then fused as $Y_{NF}$. The obtained luminance $Y_{NF}$ is consequently utilized as the guidance of the chrominance enhancement. Given the chrominance components of the low-light image, and $Y_{NF}$, the CAN outputs $C_{b_{CA}}$ and $C_{r_{CA}}$. Finally, we combine $Y_{NF}$,  $C_{b_{CA}}$ and $C_{r_{CA}}$, and convert them back to the RGB colorspace. 

To largely exclude influence from other factors and verify the main claim, all the sub-networks follow the shape shown in Fig. \ref{fig:full_architecture} B. There are three down-sampling layers and three up-sampling layers in each sub-network, and two convolutional layers with ReLU activations are inserted before each scaling layer. The inner channels of each block are doubled after each down-sampling block and halved after each up-sampling layer, changing between 32 and 128. All the convolution layers employ $3 \times 3$ kernel size, stride = 1 and padding = 1. All the sub-networks are with a Sigmoid activation at their tails, except for the ANSN which is merely the convolution. \emph{Note that our sub-networks are optimized individually.}

\subsection{IAN: Illumination Adjustment Network}
IAN is proposed to estimate the relative difference of illumination maps between low-light luminance $Y_{low}$ and the normal-light $Y_{high}$ counterpart. The difference map is adopted to adjust the low-light input into the normal-light and further contributes to the noise suppression. The following loss function is used for IAN: 
\begin{equation}
\begin{aligned}
      \mathcal{L}_{IA} = &\left\| Y_{low}/(\hat{L}+\epsilon) - Y_{high} \right\|^2_2  \\
      & +  \alpha \left\| W \circ \nabla \hat{L}\right\|_1  + \beta \left\| \nabla \hat{L} - \nabla Y_{low} \right\|_1,
\end{aligned}
\label{eq_ia}
\end{equation}
where $\|\cdot\|_1$ and $\|\cdot\|_2$ respectively stand for the $\ell_1$ and $\ell_2$ norms, and $\epsilon$ is a small constant for avoiding zero denominator. In addition, $\alpha$ and $\beta$ are two coefficients to balance the importance of different terms. $\hat{L} = \theta_{IAN}(Y_{low}) \to L_{low}/L_{high}$ is the predicted relative difference of illumination between $Y_{low}$ and $Y_{high}$. The second term is to constrain the illumination  to be piece-wise smooth with $W = 1/(\nabla Y_{low}+\epsilon)$ as the weight map. The last term is to preserve the similarity between $Y_{low}$ and $\hat{L}$ in the gradient domain for reducing halo artifacts. $\nabla$ designates the first-order derivative filter. Once $\hat{L}$ is acquired, it is used to generate the output of the first stage, following $Y_{IA} = Y_{low}/(\hat{L}+\epsilon)$. 

\begin{table*}[t]
  \caption{Quantitative results on LOL evaluation dataset of different methods in terms of PSNR, SSIM, NIQE and DeltaE. The subscript $C$ indicates gamma correction on the luminance towards references is conducted before evaluation. The best results are indicated in \textcolor{red}{red}, the second-best results in \textcolor{blue}{blue} and the third in \textcolor{green}{green}.}
  \label{tab:lol_eval}
    \centering
   
    \begin{tabular}{ccccccccc}
    \toprule[1pt]
      Metrics  & LIME \cite{guo2016lime} & SIRE \cite{fu2016weighted} & NPE \cite{wang2013naturalness}& RRM \cite{li2018structure}& EG \cite{jiang2021enlightengan}& Zero-DCE \cite{guo2020zero}& MSRNet \cite{shen2017msr}
      \\ \hline
  PSNR$\uparrow$  & 16.76 & 11.86 & 16.97 & 13.88 & 17.48 & 14.86 & 13.17  \\
  SSIM$\uparrow$ & 0.444 & 0.494 & 0.482 & 0.670  & 0.654 & 0.562 & 0.460  \\ 
  NIQE$\downarrow$ & 9.779 & 8.073 & 9.788 & 4.234  & 5.238 & 8.811 & 9.261  \\
  DeltaE$\downarrow$ & 21.43 & 32.62 & 21.77 & 26.18  & 19.31 & 24.56 & 30.17  \\
  PSNR$_{C}\uparrow$ & 19.14 & 20.97 & 20.59 & 20.25 & 22.48 & 21.88 & 16.71\\
  SSIM$_{C}\uparrow$ & 0.471 & 0.656 & 0.513 & 0.774 & 0.710 & 0.640 & 0.461\\
  NIQE$_{C}\downarrow$ & 8.954 & 7.321 & 8.890 & 3.944  & 4.837 & 7.889 & 9.074  \\
  DeltaE$_{C}\downarrow$ & 19.06 & 14.32 & 17.60 & 14.03  & 13.28 & 14.01 & 22.72  \\\hline
   
    Metrics & DUPE \cite{wang2019underexposed}& GLAD \cite{wang2018gladnet}& DRD \cite{DBLP:conf/bmvc/WeiWY018}& DRBL  \cite{yang2020fidelity}& KinD++ \cite{zhang2021beyond} & Bread & Bread-ME \\ \hline
    PSNR$\uparrow$ & 14.77 & 19.72 & 16.77 & 18.80 & \textcolor{green}{21.80}  & \textcolor{blue}{22.92} & \textcolor{red}{22.96} \\
    SSIM$\uparrow$ & 0.470 & 0.685 & 0.428 & \textcolor{green}{0.831} & \textcolor{blue}{0.836} & \textcolor{blue}{0.836} & \textcolor{red}{0.838} \\
    NIQE$\downarrow$ & 9.079 & 7.283 & 10.424 & \textcolor{green}{4.103} & 4.290 & \textcolor{blue}{3.950} & \textcolor{red}{3.946} \\
    DeltaE$\downarrow$ & 26.19 & 16.54 & 23.65 & 15.59  & \textcolor{blue}{11.52} & \textcolor{green}{11.54} & \textcolor{red}{11.19}  \\
    PSNR$_{C}\uparrow$ & 22.36 & 23.72 & 18.73 & 22.48 & \textcolor{green}{23.91} & \textcolor{blue}{25.98} & \textcolor{red}{26.06} \\
    SSIM$_{C}\uparrow$ & 0.594 & 0.724 & 0.448 & \textcolor{green}{0.851} & 0.847 & \textcolor{blue}{0.851} & \textcolor{red}{0.857} \\
    NIQE$_{C}\downarrow$ & 8.110 & 6.754 & 9.644 & \textcolor{green}{3.753}  & 3.901 & \textcolor{blue}{3.649} & \textcolor{red}{3.614} \\
    DeltaE$_{C}\downarrow$ & 14.77 & 12.54 & 21.49 & 11.75 & \textcolor{green}{10.08} & \textcolor{blue}{9.41} & \textcolor{red}{9.06}  \\
    \bottomrule[1pt]
    \end{tabular}
    

\end{table*}

\subsection{ANSN: Adaptive Noise Suppression Network}
\label{sec:noise_suppression}
According to Eq. (\ref{eq:y_high}), we obtain the adjusted luminance $Y_{IA} = Y_{high}+ \tilde{N}$, in which the noise in $Y_{low}$ is also amplified. Moreover, possible errors in the estimated illumination should not flow to the subsequent process. Therefore, we simulate amplified noise $\tilde{N}$ on normal-light references $Y_{high}$ without changing their illumination. Regions originally under darker light should have more intense noise than those under brighter conditions, thus relative illumination map $\hat{L}$ estimated previously is natural to be adopted to indicate noise levels. The choices of noise pattern are investigated in LLNet \cite{lore2017llnet}, which says that Gaussian and Poisson noises are competent. We take the simple Gaussian model for noise synthesis. Modified from the traditional AWGN model \cite{zhang2017beyond}, we reach the following for noise synthesis: 

\begin{equation}
\begin{aligned}
      Y_{N} = Y_{high} + \mathcal{N}(0,\mathcal{A}),
\end{aligned}
\label{eq_ns}
\end{equation}
where $\mathcal{A}$ can be viewed as an attention map in inverse proportion to $\hat{L}$.  
Moreover, the following simple loss function is used for ANSN: 
\begin{equation}
\begin{aligned}
      \mathcal{L}_{NS} = \left\| \theta_{NS}(Y_{N},\mathcal{A}) - \mathcal{N}(0,\mathcal{A})\right\|^2_2.
\end{aligned}
\label{eq_ns_obj}
\end{equation}
With the trained ANSN, the amplified noise in $Y_{IA}$ can be removed through $ Y_{NS} = Y_{IA} - \theta_{NS}(Y_{IA},\mathcal{A})$.

Though ANSN is already able to produce noise-free enhancement results in most cases, the solo usage of it may still leads to over-smoothed luminance for some samples. The reason is that the overall scale of the noise level map is inaccessible due to the differences in photographing devices. To robustly remove the noise, we develop a noise fusion module (NFM) to merge the luminance maps $\{Y_{NS_1}, Y_{NS_2}, ..., Y_{NS_k} \}$ with $k$ different denoising strengths of $\{ \mathcal{A}_1, \mathcal{A}_2, ...,  \mathcal{A}_k\}$. Moreover, the fusion process is expected to further remedy errors caused by the previous stages. To be concluded, the following learning problem is introduced: 
\begin{equation}
\begin{aligned}
       \mathcal{L}_{NF} &= \left\|Y_{NF} - Y_{high}\right\|^2_2 + \text{SSIM}(Y_{NF}, Y_{high})
\end{aligned}
\label{eq_fd}
\end{equation}
where $Y_{NF} = \theta_{NF}(Y_{NS_{1...k}},\mathcal{A}_{1...k})$, and  $\text{SSIM}(\cdot,\cdot)$ is the structural similarity loss. 

\subsection{CAN: Color Adaption Network}

Having obtained the $Y_{NS}$, the color components are expected to be adapted accordingly. Given the luminance and chrominance components of the original input $(Y_{low}, C_{b_{low}}, C_{r_{low}})$ and the reference $(Y_{high}, C_{b_{high}}, C_{r_{high}})$, the loss function for CAN is designed as follows: 
\begin{equation}
\begin{aligned}
       \mathcal{L}_{CA} &= \left\| C_{b_{CA}} -  C_{b_{high}} \right\|^2_2 + \left\| C_{r_{CA}} - C_{r_{high}} \right\|^2_2, 
\end{aligned}
\label{eq_ca}
\end{equation}
where  $(C_{b_{CA}}, C_{r_{CA}}) = \theta_{CA}(Y_{low}, C_{b_{low}}, C_{r_{low}}, Y_{high})$. Note that, following the divide-and-rule principle, we use $Y_{high}$ rather than $Y_{NF}$ during training to avoid the influence from possible errors left in previous stages. Once the network is trained, $Y_{high}$ is replaced with $Y_{NF}$ for testing.  

\begin{figure}[t]
\begin{subfigure}{0.32\linewidth}
    \includegraphics[width=1\linewidth]{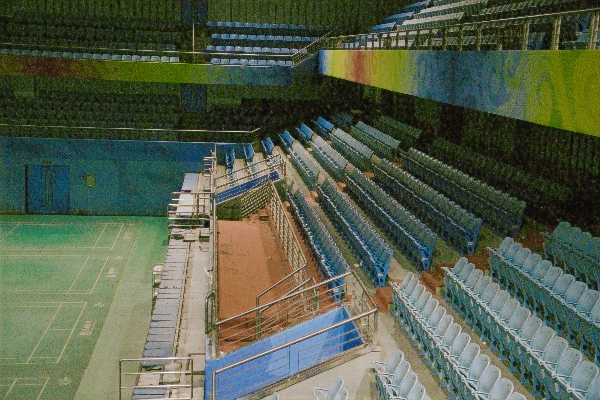}
    \subcaption*{GLAD\cite{wang2018gladnet}}
\end{subfigure}
\begin{subfigure}{0.32\linewidth}
    \includegraphics[width=1\linewidth]{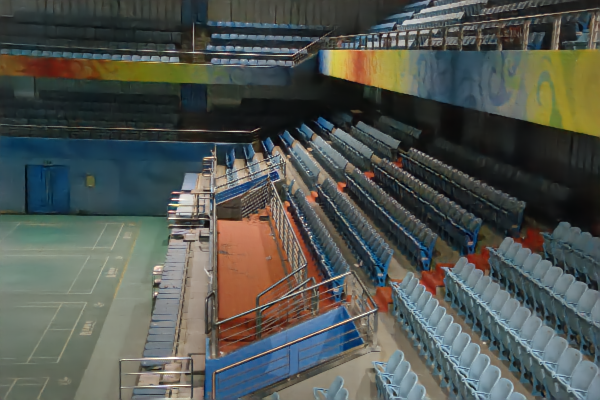}
    \subcaption*{DRBL\cite{yang2020fidelity}}
\end{subfigure}
\begin{subfigure}{0.32\linewidth}
    \includegraphics[width=1\linewidth]{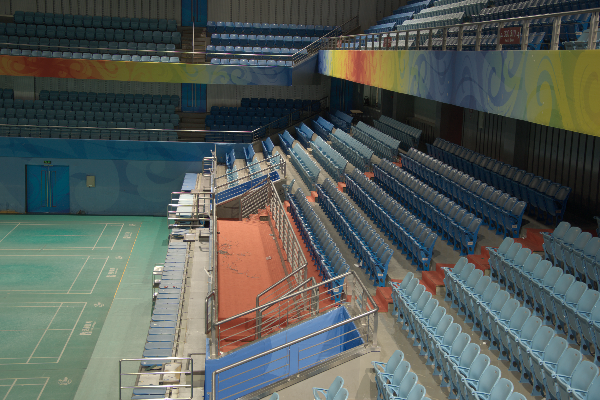}
    \subcaption*{GT (train)}
\end{subfigure} 

\begin{subfigure}{0.32\linewidth}
    \includegraphics[width=1\linewidth]{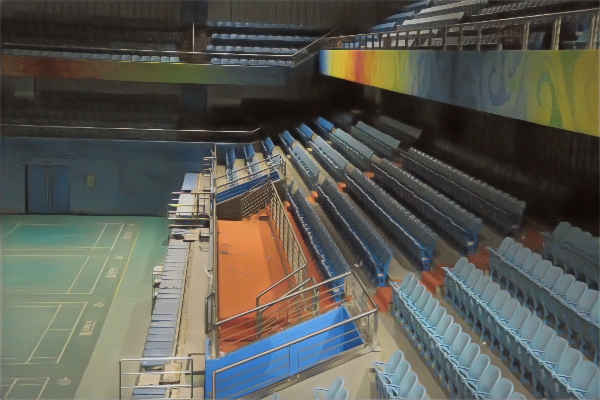}
    \subcaption*{Bread}
\end{subfigure}
\begin{subfigure}{0.32\linewidth}
    \includegraphics[width=1\linewidth]{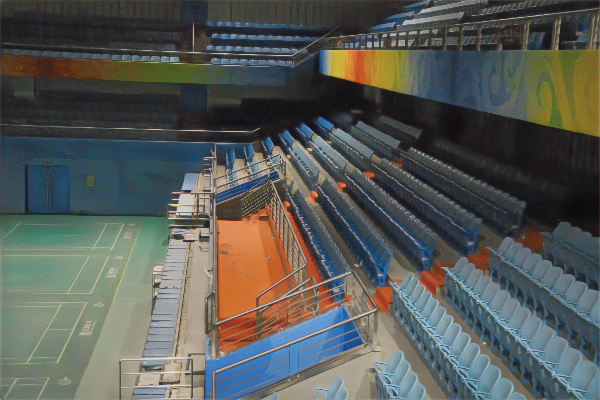}
    \subcaption*{Bread-ME}
\end{subfigure}
\begin{subfigure}{0.32\linewidth}
    \includegraphics[width=1\linewidth]{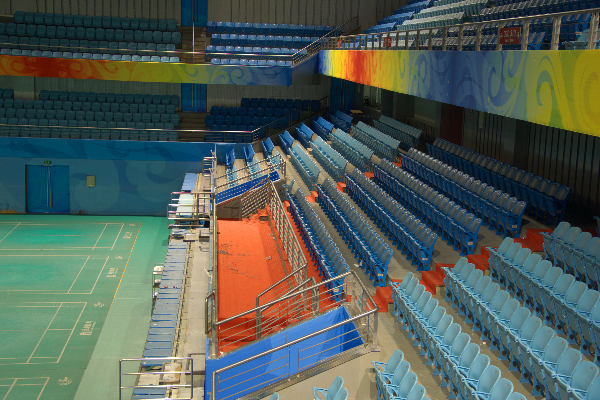}
    \subcaption*{GT (eval)}
\end{subfigure}

\caption{The ground truth of training data in the LOL dataset typically has less vivid color than that of evaluation. } 
\label{fig:lighter_color}
\end{figure}

Considering the low-saturation problem of the references in the LOL \cite{DBLP:conf/bmvc/WeiWY018} training data as shown in Fig. \ref{fig:lighter_color}, we can alternatively introduce multi-exposure data for the training of CAN, which often covers a wide range of exposure and aplenty color patterns. The following learning problem is desired: 
\begin{equation}
\begin{aligned}
       (C_{b_{ME}}, C_{r_{ME}}) = \theta_{ME}(Y_{e1}, C_{b_{e1}}, C_{r_{e1}}, Y_{e2}).
\end{aligned}
\label{eq_mef}
\end{equation}
Image pairs under different exposures, denoted as $(Y_{e1}, \\ C_{b_{e1}}, C_{r_{e1}})$ and $(Y_{e2}, C_{b_{e2}}, C_{r_{e2}})$, are randomly selected from the sequences of multi-exposure photographs. The objective function of this module shares the same form with Eqn. \eqref{eq_ca}. It forces the enhanced chrominance components to fit the exposure condition with respect to a certain luminance guided by real multi-exposure data. Our approach also answers a key question that arises in the enhancement -- to what extent the enhanced color should be. In most, if not all, of the cases, we expect it to be natural, under a certain exposure. The framework with multi-exposure data introduced is denoted as Bread-ME. We show results of both Bread and Bread-ME in quantitative experiments for fair comparisons. 

\section{Experiments}

\begin{figure*}[t]
 
\foreach \mt/\app in {Input/Input, LIME/LIME\,\cite{guo2016lime}, SRIE/SRIE\,\cite{fu2016weighted}, multiscaleRetinex/MSRNet\,\cite{shen2017msr}, dupe/DUPE\,\cite{wang2019underexposed}, DRD/DRD\,\cite{DBLP:conf/bmvc/WeiWY018}, glad_plot/GLAD\,\cite{wang2018gladnet}, DRBL_plot/DRBL\,\cite{yang2020fidelity}, KinDplus_plot/KinD++\,\cite{zhang2021beyond}, EnlightenGAN_plot/EG\,\cite{jiang2021enlightengan}, ZERODCE_plot/Zero-DCE\,\cite{guo2020zero}}{
	\begin{subfigure}{0.15\linewidth}
        \includegraphics[width=1\linewidth]{figures/DICM_sota/21_\mt.png}
        \subcaption*{\app}
	\end{subfigure}
	\hfill
 }	
\begin{subfigure}{0.15\linewidth}
    \includegraphics[width=1\linewidth]{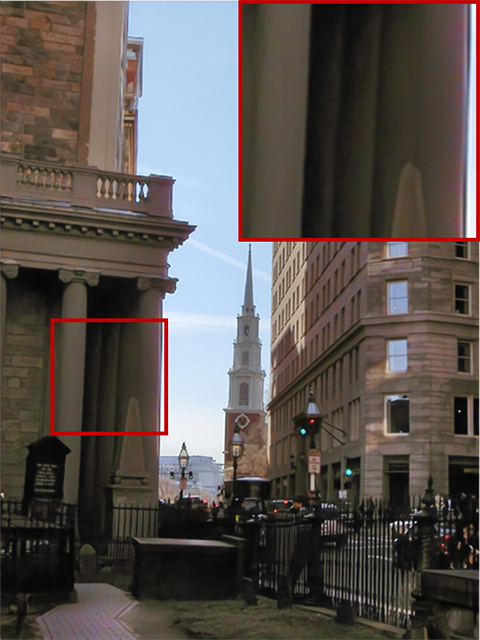}
    \subcaption*{Ours}
\end{subfigure}

\vspace{-5pt}
\caption{Visual comparison between different methods on the DICM testing dataset.  Please zoom in for details. } 
\vspace{-7pt}
\label{fig:visual_comparsion_test}
\end{figure*}

\begin{figure*}[t]
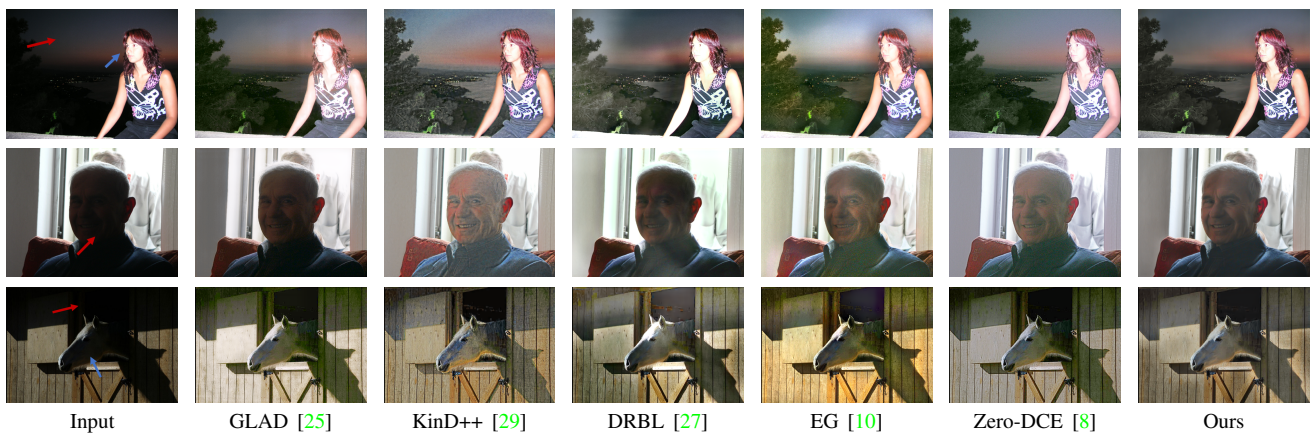

\foreach \mt/\app in {Input_PLOT/Input, glad/GLAD, KinDplus/KinD, DRBL/DRBL,  EnlightenGAN/EG, ZERODCE/Zero-DCE,BreadMEF/Ours}{
	\begin{subfigure}{0.13\linewidth}
        \includegraphics[width=1\linewidth]{figures/vv_sota/P1010062_\mt.png}
	\end{subfigure}
	\hfill
 }	

\vspace{3pt}


\foreach \mt/\app in {Input_PLOT/Input, glad/GLAD, KinDplus/KinD, DRBL/DRBL,  EnlightenGAN/EG, ZERODCE/Zero-DCE,BreadMEF/Ours}{
	\begin{subfigure}{0.13\linewidth}
        \includegraphics[width=1\linewidth]{figures/vv_sota/P1020150_\mt.png}
	\end{subfigure}
	\hfill
 }	 

\vspace{3pt}

\foreach \mt/\app in {Input_PLOT/Input, glad/GLAD\,\cite{wang2018gladnet}, KinDplus/KinD++\,\cite{zhang2021beyond}, DRBL/DRBL\,\cite{yang2020fidelity},  EnlightenGAN/EG\,\cite{jiang2021enlightengan}, ZERODCE/Zero-DCE\,\cite{guo2020zero},BreadMEF/Ours}{
	\begin{subfigure}{0.13\linewidth}
        \includegraphics[width=1\linewidth]{figures/MEFDS_sota/Horse_\mt.png}
        \subcaption*{\app}
	\end{subfigure}
	\hfill
 }	 

\vspace{-5pt}
\caption{Visual comparison between different methods on the VV (top two rows) and MEF-DS \cite{Fang2020PerceptualEF} (the last row). Please zoom in for details. } 
\vspace{-7pt}
\label{fig:visual_comparison_vv}
\end{figure*}

\subsection{Implementation Details}
All of our models are implemented in PyTorch and optimized with Adam optimizer, the parameters of which are set as $\beta_1 = 0.9$, and $\beta_2 = 0.999$. All the learning rates are fixed as $10^{-3}$, except for the finetuning of the color adaption network based on multi-exposure Data, which is set as $10^{-4}$. $\alpha$ and $\beta$ in Eqn. (\ref{eq_ia}) are set as $4.0$ and $0.5$, respectively. We set $\mathcal{A} = \textrm{exp}(-\hat{L})$, $k = 3$, $\mathcal{A}_{1,2,3} = \{0,0.05\mathcal{A},0.1\mathcal{A}\}$ for the ANSN and NFM.

We adopt the LOL dataset as the training data, which contains 485 low/normal light pairs for training and 15 for evaluation. To imitate various exposures in real-world photographs, we synthesize 8 images with different exposures for each low-light image. The magnifications of the exposures are uniformly distributed from one to making 25\% pixels over-exposed at most. For the training of the color adaption network based on multi-exposure data, we first select the SICE dataset \cite{cai2018learning} for training, then append the original 485 pairs of LOL dataset into the training data for finetuning. 

\subsection{Performance Evaluation}
To verify the effectiveness of our proposed method, several public datasets are used for evaluation, including LOL \cite{DBLP:conf/bmvc/WeiWY018}, DICM \cite{lee2013contrast}, NPE \cite{wang2013naturalness} and VV\footnote{https://sites.google.com/site/vonikakis/datasets}. Representative state-of-the-art methods, including LIME \cite{guo2016lime}, SIRE \cite{fu2016weighted}, NPE \cite{wang2013naturalness}, RRM \cite{li2018structure}, MSRNet \cite{shen2017msr}, DUPE \cite{wang2019underexposed}, GLAD \cite{wang2018gladnet}, DRD \cite{DBLP:conf/bmvc/WeiWY018}, DRBL \cite{yang2020fidelity}, KinD \cite{zhang2019kindling}, EG \cite{jiang2021enlightengan} and Zero-DCE \cite{guo2020zero} are adopted for comparisons. Image quality assessment metrics, including PSNR, SSIM, NIQE \cite{Mittal2013MakingA}, DeltaE \cite{Sharma2005TheCC} and LOE \cite{Wang2013NaturalnessPE}, are employed to measure these methods.

The quantitative comparison between Bread(-ME) and other competitors on the LOL evaluation dataset is reported in Table \ref{tab:lol_eval}. Our method outperforms other state-of-the-art alternatives by a noticeable margin, on both of the reference and no-reference metrics, demonstrating the efficacy of our proposed Bread framework.
Note that the absolute brightness is inaccessible during evaluation, which may be unfair to those non-data-driven methods and may interfere the assessment of the fidelity in details, thus we also provide a version of metrics besides the original, the predicted luminance of which is aligned to its ground truth by simple Gamma correction. Visual comparisons on several samples from the LOL evaluation and testing datasets are depicted in Figs. \ref{fig:visual_comparsion_eval} and \ref{fig:visual_comparsion_test}, respectively. The results by our method achieve remarkably higher quality with noise well-suppressed and less artifacts, while irregular illumination, noise residual, and texture/color distortion exist in the results of other methods. Moreover, we can see from Fig. \ref{fig:visual_comparsion_eval} that multi-exposure-data-based color adapter rectifies the greenish hue, producing more realistic tone. Also, we provide non-reference results in Table \ref{tab:test_niqe}. For fair comparison, the methods list in the table are all data-driven and trained on the LOL dataset. Our framework occupies the leading position in terms of both two non-reference metrics on three benchmark datasets. Please note that, NIQE does not monotonously change following the quantity of visual noises, which means that a more visual-pleasing noise-free texture can also cause a high NIQE value. We will continue this discussion in the supplementary materials. Additionally, we provide more visual results to further illustrate the effectiveness of the noise suppression and the fidelity of the color adaption in Fig. \ref{fig:visual_comparison_vv}. 


\begin{table}[t]
  \caption{Quantitative comparison on the DICM, NPE and VV datasets. All the competitors are trained on the LOL dataset.} 
  \label{tab:test_niqe}
    \centering
    \resizebox{0.47\textwidth}{!}{
        \begin{tabular}{ccccccccc}
        \toprule[1pt]
  Datasets  & Metrics & GLAD & DRD & DRBL & KinD++ & Bread & Bread-ME  \\ \hline
\multirow{2}{*}{DICM (44)} & NIQE$\downarrow$  & 3.0875 & 4.7120 & 3.2784 & \textcolor{red}{2.8584}  & \textcolor{green}{3.0893} & \textcolor{blue}{3.0869} \\
 & LOE$\downarrow$ & \textcolor{red}{240.71} & 608.70 & 660.10 & 535.34 & \textcolor{green}{400.07} & \textcolor{blue}{388.85} \\
\multirow{2}{*}{NPE (8)} & NIQE$\downarrow$  & 3.6143 & 4.0676 & \textcolor{green}{3.5843} & 3.9101 &\textcolor{blue}{3.5103} & \textcolor{red}{3.4596}  \\ 
 & LOE$\downarrow$ & \textcolor{red}{211.27} & 741.78 & 814.82 & 494.38 & \textcolor{green}{398.57} & \textcolor{blue}{392.72} \\
 
\multirow{2}{*}{VV (24)} & NIQE$\downarrow$  & 4.4112 & 4.1508 & \textcolor{red}{3.3770} & 3.8084  & \textcolor{green}{3.6770} & \textcolor{blue}{3.6710}  \\ 
 & LOE$\downarrow$ & \textcolor{red}{199.57} & 522.04 & 648.67 & \textcolor{blue}{300.57} & 339.56 & \textcolor{green}{302.26}  \\ \hline
 - & Size(MB)$\downarrow$  & 10.90 & \textcolor{green}{9.05} & 20.10 & 34.9 & \textcolor{blue}{8.21} & \textcolor{red}{8.21}  \\ 
 \bottomrule[1pt]
 \end{tabular}
 }
\end{table}
\begin{figure}[t]
\begin{subfigure}{0.32\linewidth}
    \includegraphics[width=1\linewidth]{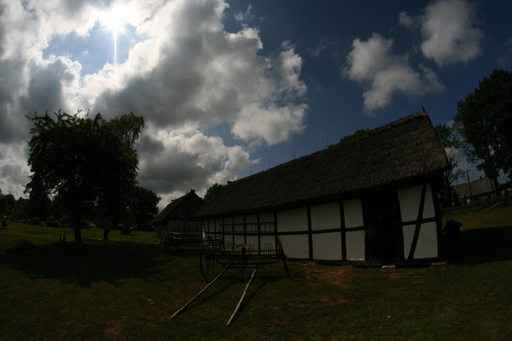}
\end{subfigure}
\hfill
\begin{subfigure}{0.32\linewidth}
    \includegraphics[width=1\linewidth]{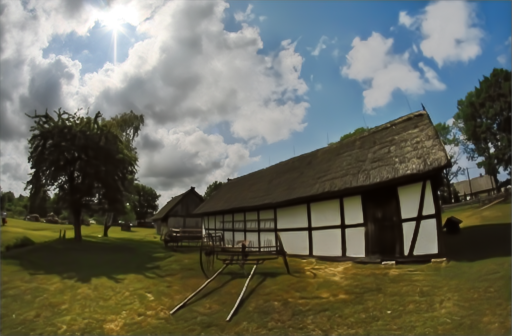}
\end{subfigure}
\hfill
\begin{subfigure}{0.32\linewidth}
    \includegraphics[width=1\linewidth]{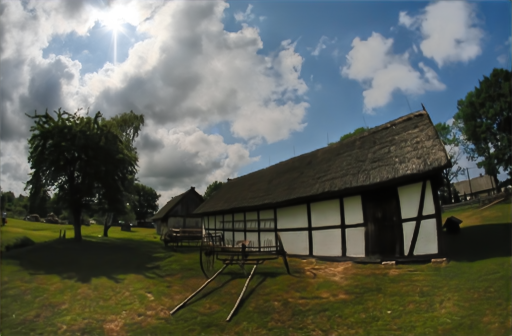}
\end{subfigure}
\hfill

\vspace{3pt}

\begin{subfigure}{0.32\linewidth}
    \includegraphics[width=1\linewidth]{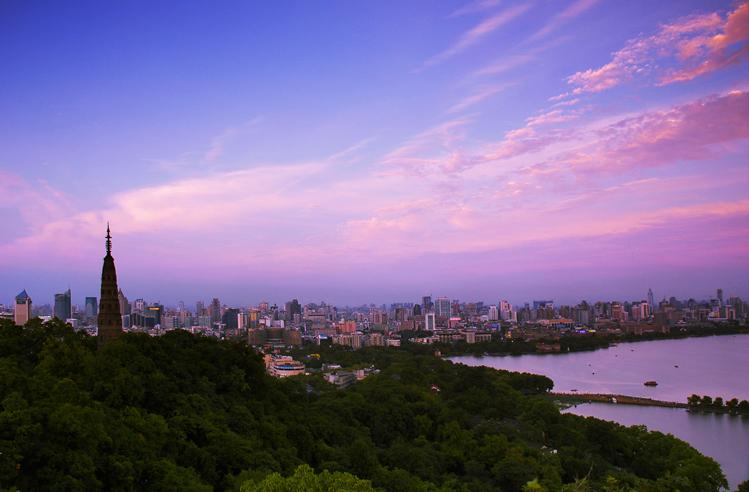}
    \subcaption*{Input}
\end{subfigure}
\hfill
\begin{subfigure}{0.32\linewidth}
    \includegraphics[width=1\linewidth]{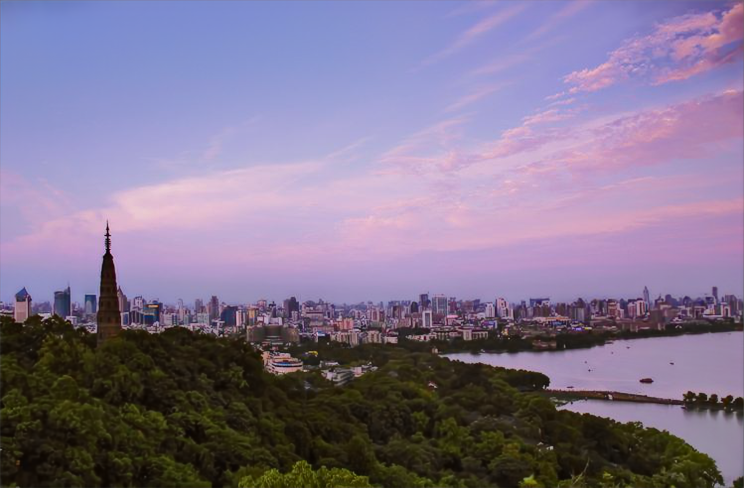}
    \subcaption*{Bread}
\end{subfigure}
\hfill
\begin{subfigure}{0.32\linewidth}
    \includegraphics[width=1\linewidth]{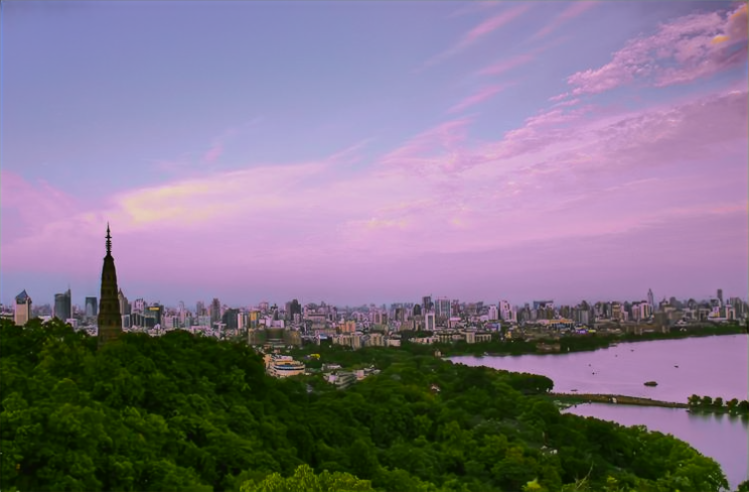}
    \subcaption*{Bread-ME}
\end{subfigure}
\hfill

\vspace{-5pt}
\caption{Visual comparison between color adaption without and with multi-exposure data used. It is obvious that the introduction of such data contributes to more vibrant color.}%
\vspace{-5pt}
\label{fig:color_fidelity}
\end{figure}

\subsection{Ablation Study}
\label{sec:ablation}

As shown in Table \ref{tab:ablation}, to evaluate the effectiveness of different settings in our framework, we conduct ablation studies, including removing the denoising process, the noise fusion module and the IA-NS separation setting, respectively. We further evaluate different noise synthesis models, including fixed Gaussian noise (noise level of 25, denoted as FGN), Poisson noise (PN) and our spatially-correlated Gaussian noise (Bread). To validate the necessity of the usage of multi-exposure data, please compare the results of  Bread and Bread-ME. The visual comparison is also shown in Fig. \ref{fig:color_fidelity}, which illustrates that the color fidelity meets significant improvement from the data. Several key questions are highlighted as follows: 

\noindent \emph{The necessity of NFM}: We can see that without the NFM, the framework encounters an obvious degradation for the non-blind metrics, because the errors arising in IAN and ANSN are left untreated. 

\noindent \emph{The necessity of IA-NS separation}: For directly learning the restoration without IA-NS separation, we use $Y_{IA}$ and $Y_{high}$ pairs and keep the same architecture of the noise suppression network. The gap between this setting and the noise synthesis approaches emphasizes the importance of the separation strategy.

\noindent \emph{The choice of noise synthesis settings}: We synthesize noisy images polluted by Poisson noise through applying the estimated low illumination map to the normal-light images and then contaminate them through the Poisson process. We then remove its illumination from the corrupted low-light images to simulate the adjusted images with amplified noise. For the fixed Gaussian results, we simply apply the AWGN on normal-light images. Unfortunately, the strength of Poisson and fixed-Gaussian denoiser is not adjustable, which means further fusion is futile, thus we adopt the spatial-correlated Gaussian as our first option. More details will be discussed in supplementary. 


\begin{table}[t]
  \caption{Ablation study on different configurations. }
  \label{tab:ablation}
    \centering
    \resizebox{0.47\textwidth}{!}{
    \begin{tabular}{lcccc}
    \toprule[1pt]
         & PSNR$\uparrow$ & SSIM$\uparrow$ & PSNR$_{C}\uparrow$ & SSIM$_{C}\uparrow$  \\ \hline
       \emph{w/o} DN & 16.91 & 0.586 & 23.33 & 0.623 \\
       \emph{w/o} NFM & 17.08 & 0.721 & 23.72 & 0.816 \\
       \emph{w/o} IA-NS Sep. & 18.69 & 0.785 & 24.58 & 0.829   \\
       \emph{w/} FGN & 21.40 & 0.799 & 25.26 & 0.821  \\ 
       \emph{w/} PN & 22.44 & 0.831 & 25.84 & 0.849   \\ 
       Bread & 22.92 & 0.836 & 25.98 &  0.851 \\
       Bread-ME & \textbf{22.96} & \textbf{0.838} & \textbf{26.06} &  \textbf{0.857} \\
    \bottomrule[1pt]
    \end{tabular}
    }
    \vspace{-10pt}
\end{table}

\section{Conclusion}
This work discussed the mixture of multi-degradation in low-light images, which increases the training difficulty and limits the enhancement quality of previous methods. To disentangle the complex degradation, the colorspace of an image is first converted from the RGB into a luminance-chrominance one, \textit{i.e.}, YCbCr, from a texture-color perspective. By doing so, the main pressure of image brightening and denoising goes to the luminance component $Y$, while the chrominance components $C_b$ and $C_r$ respond to color correction, having the enhanced $Y$ as guidance. Regarding different specific illnesses, the sub-networks including IAN, ANSN, and CAN are customized and trained individually, all of which follow the simple U-shaped net. \emph{Our designs make the training environment for each sub-network specific to one simple degradation}, the effect of which has been validated by the experiments. It is positive that such a divide-and-rule principle with texture-color decomposition can be applied to other enhancement and restoration tasks like dehazing and underwater image enhancement. 

{\small
\bibliographystyle{cvpr22}
\bibliography{cvpr22}
}

\end{document}